\documentclass[sigconf]{acmart}

\usepackage{times}
\usepackage{helvet}
\usepackage{courier}
\usepackage[english]{babel}
\usepackage[utf8x]{inputenc}

\usepackage{amsmath}
\usepackage{amsfonts}

\usepackage{graphicx}
\usepackage[colorinlistoftodos]{todonotes}
\usepackage[english]{babel}
\usepackage[utf8x]{inputenc}
\usepackage[T1]{fontenc}
\usepackage{amssymb}
\usepackage{subcaption}
\usepackage{flexisym}
\usepackage{xcolor}

\usepackage{algpseudocode}
\usepackage{algorithm}
\usepackage{algorithmicx}

\usepackage{hyperref}

\usepackage{amsmath}
\usepackage{graphicx}
\usepackage[colorinlistoftodos]{todonotes}
\DeclareMathOperator{\E}{\mathbb{E}}

\def\BibTeX{{\rm B\kern-.05em{\sc i\kern-.025em b}\kern-.08emT\kern-.1667em\lower.7ex\hbox{E}\kern-.125emX}}
 
\copyrightyear{2019}
\acmYear{2019}
\setcopyright{acmlicensed}
\acmConference[KDD 2019]{KDD 2019 Workshop}{Aug 04--08, 2019}{Alaska}
\acmBooktitle{KDD 2019 Workshop}

\begin{document}

\title{Towards resilient machine learning for ransomware detection}
\author{Li Chen, Chih-Yuan Yang, Anindya Paul, Ravi Sahita}
\affiliation{Security and Privacy Research, Intel Labs, Hillsboro, OR 97124}


%

\begin{abstract}
There has been a surge of interest in using machine learning (ML) to automatically detect malware through their dynamic behaviors. These approaches have achieved significant improvement in detection rates and lower false positive rates at large scale compared with traditional malware analysis methods. ML in threat detection has demonstrated to be a good cop to guard platform security. However it is imperative to evaluate - is ML-powered security resilient enough? 

In this paper, we juxtapose the resiliency and trustworthiness of ML algorithms for security, via a case study of evaluating the resiliency of ransomware detection via the generative adversarial network (GAN). In this case study, we propose to use GAN to automatically produce dynamic features that exhibit generalized malicious behaviors that can reduce the efficacy of black-box ransomware classifiers. We examine the quality of the GAN-generated samples by comparing the statistical similarity of these samples to real ransomware and benign software. Further we investigate the latent subspace where the GAN-generated samples lie and explore reasons why such samples cause a certain class of ransomware classifiers to degrade in performance. Our focus is to emphasize necessary defense improvement in ML-based approaches for ransomware detection before deployment in the wild. Our results and discoveries should pose relevant questions for defenders such as how ML models can be made more resilient for robust enforcement of security objectives.

\end{abstract}

%
%

%
\keywords{Dynamic ransomware detection, generative adversarial net, adversarial quality metric.}

\maketitle
\section{Introduction}

Ransomware is a type of malicious software (malware), which hijacks and blocks victim's data or machine until a monetary ransom is paid. Its life cycle consists of six phases \cite{mcafeelabs2016}: i). Distribution: the ransomware arrives at victim’s machine by an email attachment, a drive-by download or a code dropper; ii) Infection: the ransomware installs itself to survive a reboot and disables shadow copies or anti-virus processes; iii). Communication: the ransomware contacts its Command and Control (C\&C) server for the encryption key; iv) Preparation: the ransomware scans the user’s files, usually pdf, docx, jpg files; v).  Encryption: the ransomware encrypts the selected user files; and finally vi) Extortion: a “ransom note”, asking for payment, is displayed to the user. After the ransom is paid, instructions to receive the decryption key will be sent to the user. 

There are two main categories of ransomware based on attack approaches: the locker-ransomware and the crypto-ransomware \cite{cabaj2018software, gomez2018r}. 
The locker-ransomware locks the victim's computer without the encryption. The crypto-ransomware encrypts victim’s files which are very difficult to revert. The quick solution is to pay the extortion and hope the given key can truly decrypt the data. Thus the crypto-ransomware remains a notorious security issue today. In our case study, we focus on cryto-ransomware. 



The popularity of Internet and untraceable payment methods and availability of software development tools makes ransomware an feasible weapon for remote adversaries \cite{al2018ransomware}.
In recent years, ransomware has posed increasingly major threats. Since 2017, ransomware attacks have increased over 59\% yearly with 35\% growth in Q4 2017 alone. Although the trend of devising new ransomware has declined in 2018, the occurrence of ransomware attacks is still rising~\cite{mcafee032018, mcafee092018}. 

Dynamic analysis on malware can reveal true malicious intentions by executing malware in a contained environment. 
Recent research have found behavior analysis via analyzing API calls, registry accesses, I/O activities or network traffic can be effective for ransomware detection\cite{cabaj2018software, hampton2018ransomware, scaife2016cryptolock, kharraz2016unveil, morato2018ransomware, gomez2018r, continella2016shieldfs}. 

Faced with a tsunami of malware attacks, the security industry are employing machine learning (ML) to automatically detect threats and enhance platform security. 
Their confidence in ML is not ungrounded.
ML algorithms have demonstrated state-of-the-art performance in the field of Computer Vision (CV), Natural language Processing (NLP), Automatic Speech Recognition (ASR). 
The success of ML has generated huge interest of applications on platform security domains for automated malware detection\cite{rieck2011automatic, chen2018henet, shamili2010malware, chen2017semi}. 
Particularly for ransomware detection, algorithms such as naive Bayes, support vector machine, random forest, logistic regression have shown good classification efficacy \cite{narudin2016evaluation, alhawi2018leveraging, ransomware_rsa2017}.
Shallow or deep neural networks also demonstrated high effectiveness at ransomware detection \cite{vinayakumar2017evaluating, aragorn2016deep, chen2017deep}. 

Recent research take advantage of opaqueness of NN algorithms and generate subliminal perturbed input examples which have shown to evade ML based detection.
These types of emerging attacks, where an adversary can control the decision of the ML model by small input perturbations, expose a broad attack surface. 
Although most of the Adversarial Machine Learning (AML) publications \cite{carlini2017adversarial, biggio2018wild, dalvi2004adversarial, ilyas2018black, carlini2018audio} focus on misclassification on CV and ASR domains, the proliferation of adversarial examples are spreading to generate sophisticated adversarial malware. These examples perform real-time evasive attack by camouflaging malicious behavior to a legitimate software while keeping maliciousness intact and fooling detection during run-time. For example, AVPASS\cite{jungavpass-bh} generates potent variations of existing Android malware by querying and inferring features used by malware detection systems.
Additionally recent research have shown promise of ML-based approaches to thwart ransomware attack on user systems \cite{continella2016shieldfs, sgandurra2016automated}. 

The malicious use of ML motivates us to properly study the adversarial attack threat models and investigate the robustness and vulnerability of ML-powered security defense systems. In this paper, we present a case study on using deep learning to automatically bypass ML-powered dynamic ransomware detection systems. We propose a framework based on generative adversarial network \cite{goodfellow2014generative} to generate dynamic ransomware behaviors and a set of adversarial quality metrics to justify the generated samples indeed persist maliciousness. We discover that most of the selected highly effective ransomware classifiers fail to detect the adversary-generated ransomware, indicating a broad attack surface for ML-powered security systems. We thoroughly examine the latent feature space to understand where the adversarial examples lie. We believe that our proposed framework is useful for the defender system to incorporate and minimize their detection algorithms' blind spots. Our case study examines the roles of ML as both a good cop and a bad cop for platform security. 

The goal of our paper is to provide a framework to understand the resiliency of ransomware detectors. We do not enable a true attack on user system. 
As demonstrated in this paper, we advocate that a defender should fortify their ML models for ransomware detection via adversarial studies.



Our contributions are summarized as follows:
\begin{enumerate}
    \item Although generative adversarial network (GAN) has been used to generate fake samples to resemble the true data distribution, our framework is the first one to study ML resiliency via GAN to automatically generate dynamic ransomware behaviors. Although our experiments illustrated that ML models are highly effective in combating real-world ransomware threats and can achieve high classification accuracy up to 99\% accuracy with extremely low false positive rate, our results show that such ML models fail to detect the GAN-generated adversarial samples. To stabilize training and achieve convergence, we utilize data segmentation techniques and auxiliary conditional GAN architecture.

    \item We propose a set of adversarial quality metrics to validate the generated adversarial ransomware and demonstrate the GAN-generated samples via our framework maintain maliciousness verified by such metrics. Although our ML classifiers misclassify these adversarial samples as benign, the adversarial samples are statistically much closer to real ransomware samples. 
    
    \item We emphasize that robustness against adversarial samples is an equally important metric in addition to accuracy, false positive rate, true positive rate, $F_1$ score to thoroughly evaluate ransomware detection scheme before deployment. In our experiment, only one of the seven models has the strongest resiliency on the GAN-generated samples, indicating a broad adversarial attack surface of ML algorithms. On the other hand, our experiments provide guidance for security practitioners to develop resilient ML algorithms proven to defend against adversarial attacks.
    

    
    
    
    \item We study the reasons why the highly effective models are susceptible by properly investigating in the latent feature space and provide understanding of the blind spots of these models. We present our learning to generate awareness to the security community that adversarial threat models need to be properly evaluated before deploying ML models to detect malware attacks.  
    
\end{enumerate}

The rest of the paper is organized as follows: Sec \ref{sec:related_work} briefly provides the background on ransomware analysis, adversarial machine learning and generative adversarial network. Sec. \ref{sec:data} describes system architecture, data collection and pre-processing. Sec. \ref{sec:method} presents our proposed framework and adversarial quality assessment procedure. Sec. \ref{sec:exp} illustrates experimental results on our dataset.  

\section{Background and Related Work} \label{sec:related_work}
\subsection{Ransomware Detection}

Cabaj et al. \cite{cabaj2018software} use HTTP message sequences and content sizes to detect ransomware. Morato et al. \cite{morato2018ransomware}  analyzed file sharing traffic for ransomware early detection. Scaife et al. \cite{scaife2016cryptolock} provide an early detection system by monitoring user data changes including the file entropy and similarity changes, the file type changes, file deletion and file type funneling. The honeyfiles-based R-Locker system is in \cite{gomez2018r} to trap and block ransomware operations. When ransomware scans user’s file system and accesses pre-installed decoy files, the R-Locker service is triggered to apply countermeasures. The “Unveil” system introduced in \cite{kharraz2016unveil} can detect crypto-ransomware via the I/O access patterns. A Windows kernel I/O driver is developed to collect I/O operations and  buffer entropy. It provides an early detection capability on a zero-day ransomware. 
Continella et al. create ShieldFS \cite{continella2016shieldfs}, a  custom kernel driver that collects and performs analysis of low-level file-system activity to classify ransomware activity at runtime using a multi-tier hierarchical decision tree based process monitoring model. ShieldFS also integrates file back-up to its ransomware detection system so it can able to recover files from a trusted secure storage after confirming malicious activity. Sgandurra et al. \cite{SgandurraMML16} proposed "EldeRan" which dynamically analyzes Windows API calls, registry key operations, file system operations, directory operations and so on in a sandboxed environment, selects relevant features and finally applies a logistic regression classifier to determine whether an application is ransomware or benignware. In contrast to monitoring system executions proposed in "EldeRan", we have focused on collecting changes in user file events which demonstrated early indication of ransomware activity, helped processing and storing limited data and finally worked under very limited computational budget in order not to interfere with users regular computational needs. Scaife et al. proposed CryptoDrop \cite{Scaife2016DCS}, an early stage ransomware detection system which made use of file event changes rather than program execution inspection through API call monitoring. Although they have used different ransomware behavioral indicators compared to our file events, they have reached the same conclusion as ours that an union of indicators is a more effective approach in ransomware detection than any of those alone. 

\subsection{Adversarial Machine Learning}
The first adversarial machine learning attack is used against spam filtering by generating adversarial text without affecting content readability \cite{Dalvi1014066}. The topic got significant attention in the security community when Szegady et al. \cite{SzegedyZSBEGF13} fool a DNN based image recognition classifier by adding low-intensity perturbations to the input image which looks indistinguishable to human eyes. 
Adversarial attacks on CV typically add small human imperceptible perturbations to the original images and have shown to drastically alter the ML boundary decisions \cite{DBLP:journals/corr/GoodfellowSS14}, \cite{DBLP:journals/corr/KurakinGB16}, \cite{DBLP:conf/eurosp/PapernotMJFCS16}, \cite{DBLP:conf/cvpr/Moosavi-Dezfooli16}, \cite{DBLP:conf/sp/Carlini017}. 
Beyond CV, \cite{DBLP:journals/corr/abs-1801-01944} generate adversarial speech to change the output of a speech-to-text transcription engine. Adversarial malware are created to bypass ML detection while keeping maliciousness of the software intact~\cite{jungavpass-bh}. 

Defense techniques including pre-processing via JPEG compression \cite{DBLP:conf/kdd/DasSCHLCKC18, das2017keeping}, feature squeezing \cite{DBLP:conf/ndss/Xu0Q18}, architecture via regularization \cite{DBLP:journals/corr/abs-1803-06373}, adversarial training \cite{DBLP:journals/corr/MadryMSTV17}, neural fingerprinting \cite{DBLP:journals/corr/abs-1803-03870} have exhibited success to mitigate the proliferating adversarial machine learning attacks.

\subsection{Generative Adversarial Network}

The first generative adversarial network (GAN) ever introduced is a fully connected neural network architecture for both the discriminator and the generator~\cite{goodfellow2014generative}. Ever since, abundant GAN variants are proposed. The Deep Convolutional GAN (DCGAN) \cite{radford2015unsupervised} proposes using strided convolutions instead of fully connected multi-layer perceptrons and feature normalization to stabilize training and dealing with the poor weight initialization problem. The Conditional GAN (CGAN) \cite{mirza2014conditional} adds conditional setting to the generator and the discriminator by making both neural networks class-conditional. It has advantages to better represent multi-modal data generation. The Laplacian Pyramid GAN (LPGAN) \cite{denton2015deep} produces high quality generated images and uses multiple generators and discriminators in its architecture. It downsamples the input images, and during backpropagation, injects noise generated by a conditional GAN and then upsamples the images. Auxillary Classifier GAN (ACGAN) \cite{odena2016conditional} improves the training of GAN by adding more structure to the GAN's latent space along with a specialized cost function. Wasserstein GAN (WGAN) \cite{arjovsky2017wasserstein} uses Wasserstein distance as the loss function to efficiently approximates the Earth Mover distance and significantly reduces the mode dropping phenomenon. 

Generative adversarial network has been used in creating adversarial examples to fool ML. \cite{xiao2018generating} trains a conditional GAN algorithm, AdvGAN, to generate  perceptually similar adversarial input images to attack state-of-the-art defense methods \cite{madry2017towards}. \cite{hu2017generating} uses a method to generate adversarial malware samples using MalGAN to attack state-of-the-art black-box ML detection algorithms. \cite{anderson2016deepdga} demonstrates creation of domain generation malware instances (DGAs) using GAN to bypass modern DGA ML classifiers such as random forest. \cite{bhaskara2018emulating} proposes using GAN to model the malicious behaviors and generate synthetic malware representation which is trained with existing malware samples for effective zero-day threat prevention on ML detectors. \cite{kim2018zero} proposes tDCGAN using unsupervised deep auto-encoding technique to generate new malware variants based on raw codes and modified features. 

\section{Ransomware Data Description} \label{sec:data}
\subsection{Data Collection and Description}

\begin{figure}[t!]
\centering
\includegraphics[width=0.35\textwidth]{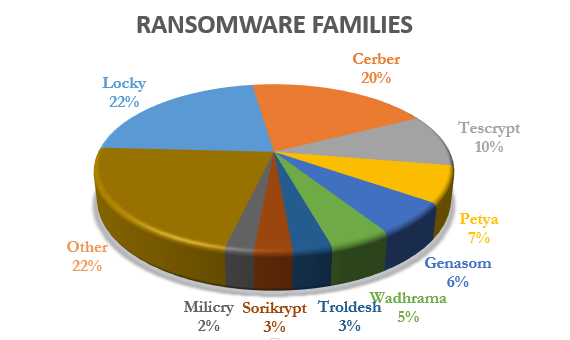}
\caption{\label{fig:families} The major ransomware families in our dataset. The distribution is based on the labels provided by Microsoft in VirusTotal.} 
\end{figure}

In our analysis, the ransomware samples are downloaded from VirusTotal, where we collect submitted ransomware between late 2017 to early 2018 based on tags from Microsoft and Kaspersky. The dataset contains various of ransomware with nine major families including Locky: a Microsoft Office macro based ransomware, and Cerber: a product of ransomware-as-a-service. The ransomware family distribution is seen in Figure ~\ref{fig:families}.

\begin{figure}[t!]
\centering
\includegraphics[width=0.45\textwidth]{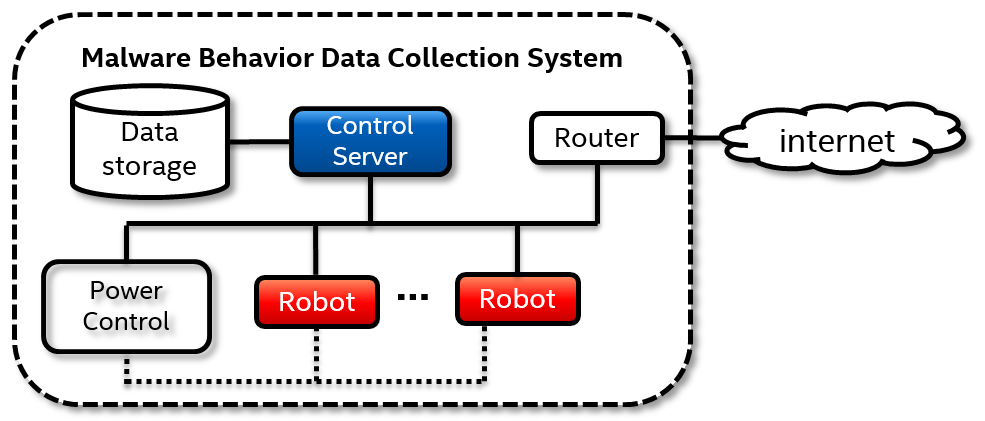}
\caption{\label{fig:method} A diagram of behavior data collection system. The robot, a Windows PC, will execute a binary downloaded from the Control server. The execution log is uploaded to the Data storage. Power Control can shut down robots if needed.}
\label{fig:sandbox}
\end{figure}

The samples are executed in a regular Windows system as shown in Figure ~\ref{fig:sandbox}. The dynamic behaviors are collected via the .Net framework FileSystemWatcher (FSW) API. The callback functions bound with FSW are triggered for all file I/O operations. We collect the low-level I/O activity patterns and calculate the normalized Shannon entropy of the targeted files \cite{scaife2016cryptolock}. To catch evasive ransomware, a user activity simulation program is executed to emulate mouse clicks and key strokes. To mimic an active desktop environment, a Notepad and Office Word applications are launched before and during ransomware execution. The benign data is collected manually from installing and executing approximately a hundred applications from various categories such as office suite, browsers and file compression applications. The idle I/O activities of benign Windows system are collected for a few months from regular backups, updates, anti-virus applications and so on.

Each data collector robot, as seen in Figure ~\ref{fig:sandbox}, is pre-installed with several user files such as Windows Office, text or multimedia files. These files are designed to be the target of ransomware and used as decoy files to filter active ransomware samples. If these files are modified during execution, then this sample is assumed to be a “crypto-”ransomware and then collected to the malicious dataset. All behavior data are uploaded to Resilient ML platform \cite{mlplatform} for data cleansing. The names of the decoy files are appended with time stamps before ransomware execution, so each sample will see the same set of user files but with different file names. 


\subsection{Feature Mapping}

\begin{figure}[h!]
\centering
\includegraphics[width=.5\textwidth]{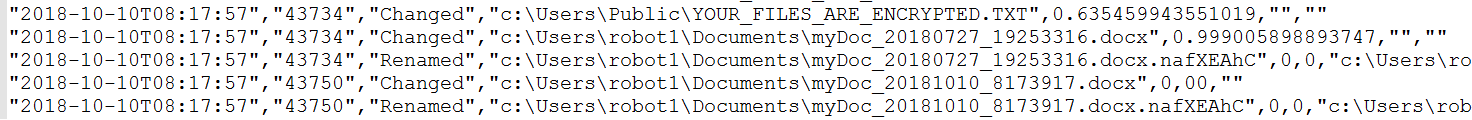}
\caption{\label{fig:method} A screen shot of dynamic execution log collected using \textit{FileSystemWatcher} (FSW).}
\label{fig:screen_of_fsw}
\end{figure}

The collected execution log via FSW contains time stamp, event name, targeted file name and file entropy, as seen in Figure \ref{fig:screen_of_fsw}. We attempt the least effort of feature processing by mapping the event combined with entropy change. The four main file actions are file delete, file create, file rename and file change. The entropy level is combined with the event of file change. Hence each execution log is represented by a sequence of events. We set the length for each sample to be 3000, so that the shorter length samples will be padded with zeros towards the beginning to match the dimension. Table \ref{tab:feature_mapping} shows the feature mapping. 
\begin{table}[h]
\center
\tiny
\begin{tabular}{| l | c |  }
\hline
  Events & Feature encoding\\ \hline
  Padding & 0 \\
  File deleted & 1 \\
  File content changed and entropy $\in [0.9, 1]$ & 2\\
  File content changed and entropy $\in [0.2, 0.4] $ & 3\\
  File content changed and entropy $\in [0, 0.2]$ & 4\\
  File created & 5\\
  File content changed and entropy $\in [0.8, 0.9]$ & 6\\
  File renamed & 7\\
  File content changed and entropy $\in [0.4, 0.6]$ & 8\\
  File content changed and entropy $\in [0.6, 0.8]$ & 9\\ \hline
\end{tabular}
\caption[Feature mapping]{Feature mapping. We attempt the least effort of feature processing, categorize the events into 9 categories and use zero for padding maintain the same length. Our features contain various I/O events and entropy change.}
\label{tab:feature_mapping}
\end{table}

The patterns from both I/O event types and entries of target file constitute our feature set. The benign programs, such as WinZip, may have \textit{change} events with high entropy, but they will not have \textbf{ as many} \textit{rename} or \textit{delete} I/O events as typical crypto-ransomware. On our collected dataset, combining various I/O event types and entropy as features is effective to detect ransomware.
\section{Synthesizing dynamic features via GAN}\label{sec:method}

GANs are mostly used in computer vision to generate images that seem real to the human eyes. Because they are typically used in CV, one can terminate the training when the generated images look like the real images. The inputs in our case study, however, are dynamic execution logs, so it is not practical to stop training GAN by merely visualizing the generated samples. Furthermore when we directly employ the typical training mechanism of GANs, mode collapsing issues constantly arise. The challenges of training an effective GAN to fool the ransomware classifier motivate us to propose a different GAN training scheme for faster convergence and better-quality sample generation. 

The principle of our proposed GAN training scheme is to segment the dynamic execution logs and leverage transfer learning to accelerate training convergence. Each execution log is segmented into $m$ subsequences and then converted 2-dimensional arrays. Then transfer learning is employed such that the parameters and neural network architectures are borrowed from existing and successfully convergent GANs used in the vision domain, while we still train from scratch on the fixed architecture. The effectiveness of employing transfer learning from computer vision to malware classification is previously demonstrated for both static and dynamic malware classification \cite{chen2018deep, chen2018henet}, but not yet for adversarial malware generation.

\subsection{Threat Model}
We assume that the adversary has knowledge to the training data, but no knowledge at all of the underlying ransomware classifiers. This is a realistic assumption since for malware detection, anti-virus vendors obtain their training samples from VirusTotal, which allows users to download binaries or hashes. 

\subsection{Training Procedure}
\begin{figure}
\centering
\includegraphics[width=0.45\textwidth]{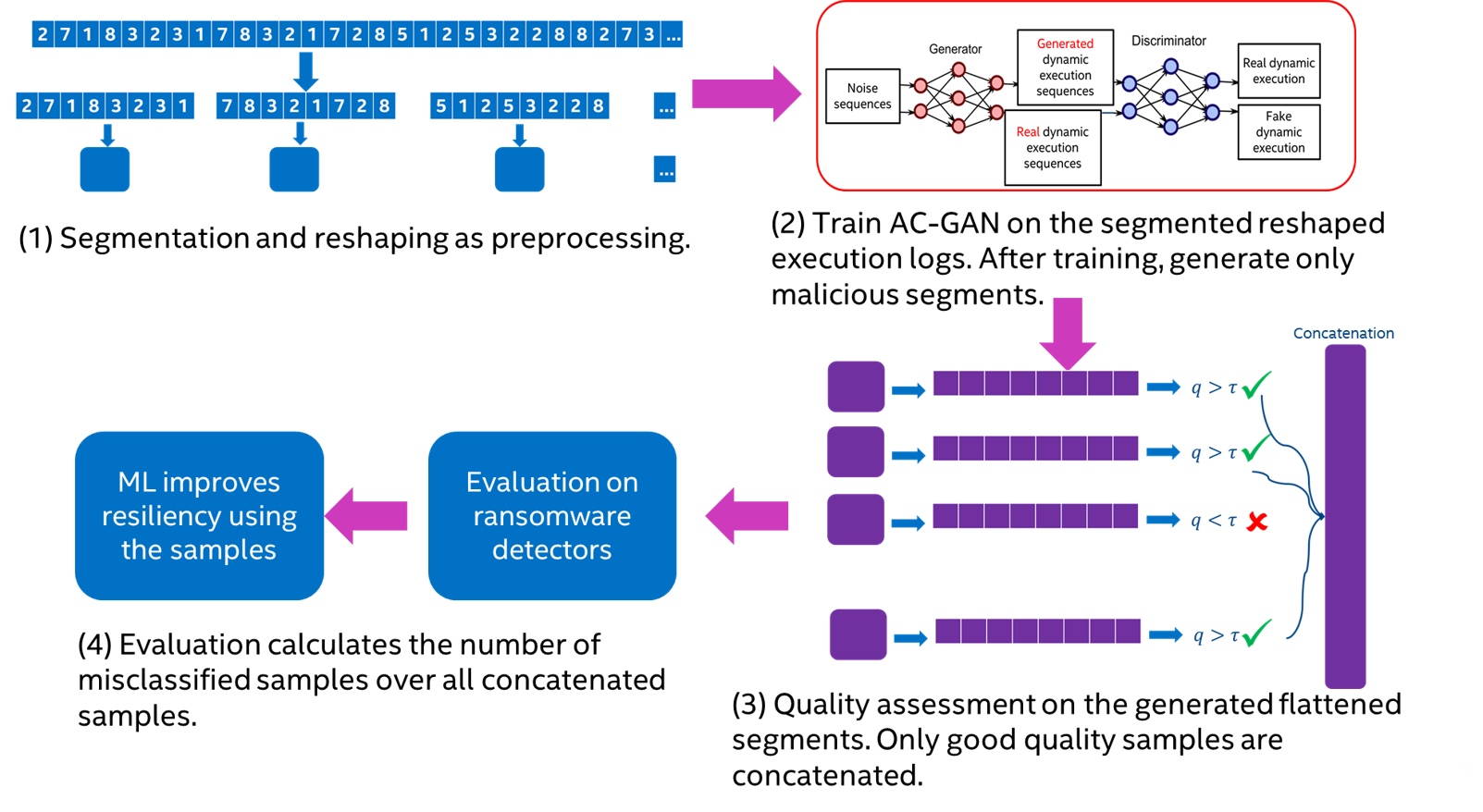}
\caption{\label{fig:overview} Overview of our proposed framework using GAN to generate dynamic ransomware features to bypass detection. }
\end{figure}

Our approach essentially consists of segmentation and reshaping as preprocessing, GAN training, quality assessment, concatenation and evaluation. An overview of our framework is seen in Figure \ref{fig:overview}.

\subsubsection{Segmentation and reshaping as preprocessing} We observe that, in our initial experiments, GAN did not converge when trained on the entire logs. This motivates us to consider training a convergent GAN on log segments. After feature mapping, we divide each training execution log into sequences each of length $784$. If the length of the execution log is not divisible by $784$, the end of the last subsequence will be padded zero. Each subsequence is then reshaped into two-dimensional square arrays of $28 \times 28$. 

We note that the convergence issue may be resolved through searching the space of neural network architectures and the parameters. However our preprocessing step enables transfer learning to borrow existing convergent GAN architectures, hence saving exhaustive search efforts while still achieving convergence. 

\subsubsection{Training}
\label{subsec:acgan}
The generative adversarial networks (GAN), first introduced in \cite{goodfellow2014generative}, are paired neural networks consisting of a generator and a discriminator, which act like two players to win a game. The generator produces samples from the generated distribution $P_G$ which is to be as close as the real data distribution $P_R$. The discriminator classifies whether the samples are generated by $P_G$ or truly sampled from $P_R$. The purpose of the generator is to fool the discriminator and the purpose of the discriminator is to separate the fake from the real. At the end of the training, the generator is supposedly and theoretically to maximize fooling the discriminator. 

\begin{figure}
\centering
\includegraphics[width=0.4\textwidth]{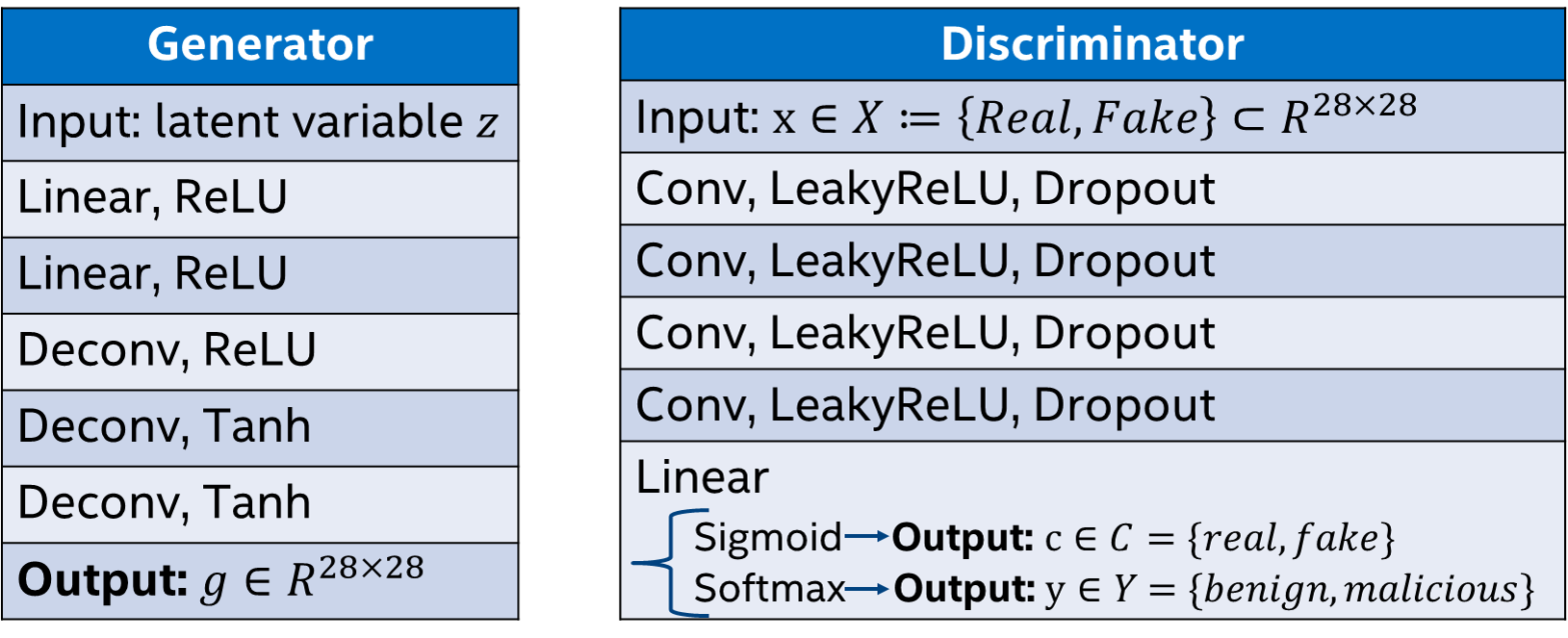}
\caption{\label{fig:ac_gan} The ACGAN architecture for generating reshaped execution segments. Left table: the architecture of the generator, where the input $z$ is latent variable and the output $g$ is a generated 2-D execution log segment. Right table: the architecture of the discriminator, where the inputs are the 2-D execution log segments, the output $y\in Y$ is predicted as benign or malicious via the auxiliary classifier, and the output $c\in C$ is predicted as real or fake. }
\end{figure}

We train an auxiliary classifier generative adversarial network (ACGAN) on the segmented two-dimensional arrays processed from the execution logs. The ACGAN architecture we employed from computer vision is shown in Figure \ref{fig:ac_gan}. Denote each real sample as $r\in \mathcal{R} \subset \mathbb{R}^{28\times28}$, where $\mathcal{R}$ is the space containing all the real segmented execution logs. The paired data are drawn from the joint distribution $(r_1, y_1), (r_2, y_2),...  (r_n, y_n)\stackrel{i.i.d}{\sim} P_{\mathcal{R},Y}$, where $y\in Y$ are the class labels with $Y=1$ being ransomware and $Y=0$ being benign. 

Denote each generated sample as $g\in \mathcal{F}$, where $\mathcal{F}$ is the space containing all fake samples and $g$ is drawn from the generated sample distribution $g\in P_G$.  Let random variable $C$ denote the label for data source where $C=1$ means the data is real and $C=0$ means the data is fake. The entire data denoted by $X$ consist of both real and fake samples, i.e., $ X = \mathcal{R}\cup \mathcal{F}$. 

We denote $z$ as the noise generated by the generator $G$, which is a function $G: (z, y) \mapsto g$.
Given the data $X$, the discriminator $D$ calculates two probabilities:  whether the data is real or fake $\mathbb{P}(C|X)$ and the class label of the sample $\mathbb{P}(Y|X)$. The loss function of AC-GAN comes into two parts: 
\begin{equation}
    L_C = \E(\log \mathbb{P}(C=1|\mathcal{R})) + \E (\log \mathbb{P}(C=0|\mathcal{F})),
\end{equation}
and
\begin{equation}
    L_Y = \E(\log \mathbb{P}(Y=y|\mathcal{R})) + \E (\log \mathbb{P}(Y=y|\mathcal{F})). 
\end{equation}
The generator is trained to maximize $L_Y - L_C$ and the discriminator is trained to minimize $L_Y + L_C$. Adding the above auxillary classifier to the discriminator in AC-GAN stabilizes training. 


Because our threat model assumes the adversary has no knowledge of the underlying classifier, the stopping criterion for training our proposed mechanism only relies on the discriminator loss. However in a white-box attack where the adversary has knowledge of the ransomware detector, the goal of the attacker is to cause the generated samples from the malicious class to be misclassified as benign. Hence we can include a third term, with respect to the ransomware detector, to the loss function as follows:
\begin{equation}
    L_{detector} = \E(\log \mathbb{P}_G( \Hat{Y} = 0 | Y=1, C = 0)).
\end{equation}

The stopping criterion for training is the loss of the discriminator. 
After training, we can generate both fake malicious samples $G_m$ and fake benign samples $G_b$. From an attacker's perspective, it is more desirable to generate malicious samples, bypass detection and increase false negative rate. Hence we focus on $G_m$ for subsequent analysis and experiments. Each generated sample is of size $28 \times 28$, so we flatten the sample to 1-dimensional segments of length $784$ and round the generated sample to the closest integer value. For abuse of notation, we denote this set as $G_m$.


\subsection{Quality Assessment on the Generated Malicious Samples}
\label{subsec:quality_assess}

Unlike in computer vision where the quality of the generated samples can be evaluated by visual inspection, evaluating the quality on dynamic execution logs requires a quantifiable metric. We propose a sample-based adversarial quality metric  $q_i$, where for each sample $i \in G_m$ 

\begin{equation}
    q_i = \frac{n_1(i)}{n_2(i)},
\end{equation}
where $n_1(i) =|\mathcal{N}_{i\cap m} | -|\mathcal{N}_{i\cap m \cap b} |  $, and $n_2(i) = |\mathcal{N}_{i \cap b} | -|\mathcal{N}_{i\cap m \cap b} |  $. Here, $|\cdot|$ denotes the cardinality, $\mathcal{N}_{i\cap m}$ is the set of matched $n$-grams between the sample $i$ and the malicious test set,  $\mathcal{N}_{i\cap b}$ is the set of matched $n$-grams between the sample $i$ and the benign test set and $\mathcal{N}_{i\cap m\cap b}$ is the set of matched $n$-grams among the sample $i$, the test malicious set and the test benign set. 
Passing the quality check means that the generated samples contain more unique malicious samples than the unique benign samples. Since the real test data was not used for training the ACGAN, the proposed metric evaluates the generalized malicious properties that may not be found from the training set. 

For a generated set $G_m$, we calculate the quality metrics for each sample and filter the samples whose quality metric is below a pre-specified threshold $\tau$. Suppose we expect to generate $K$ malicious samples and $K_0$ samples have $q < \tau$. Then we regenerate a smaller set of $G_m^{'}$, and repeat the process until we obtain $K$ desired quality samples. 

Similarly for the entire set $G_m$, we propose a batch-based adversarial quality metric $Q$ to statistically summarize the set of $q_i$ for all $i \in G_m$. The summary statistics are minimum, first quartile, median, lower quartile, minimum and outliers.

We summarize the adversarial quality assessment procedure in Algorithm \ref{alg:quality_assessment}.
\begin{algorithm}[h]
\begin{algorithmic}
\State \textbf{Input}: Generated set $ G_m$ with $|G_m|=K$ and quality threshold $\tau$   
\State \textbf{Output}: $K-K_0$; $G_{\{m, q<\tau}\}$ 
\State  Step 1: Calculate $\{q_1, \dots, q_K \}$. 
\State  Step 2: Remove samples with bad quality $q < \tau $. Denote the set of bad samples by $G_{\{m, q<\tau}\}$ where $|G_{\{m, q<\tau}\}| = K_0$. \\ 
 
 \caption{Adversarial quality assessment procedure}
 \label{alg:quality_assessment}
 \end{algorithmic}

\end{algorithm}

\subsection{Log Generation and Evaluation}
\label{subsec:eval}
The number of ways to concatenate the generated segments from $G_m$ is approximately $\lfloor m^{\frac{3000}{784}} \rfloor$. In our experiment, since all the segments in $G_m$ pass quality assessment, we can randomly concatenate the individual segments. We note that for even stronger attacks, the attacker can optimize the concatenation based on some optimization objective, and this is one of our next research steps. 

The generated malicious samples, after quality assessment in Sec \ref{subsec:quality_assess}, are fed into the ransomware classifier. The adversarial detection rate is defined as the number of correctly predicted adversarial samples divided by the total number of adversarial samples. From a defender's perspective, we can use the adversarial detection rate as another metric to quantify how resilient the malware detector is against adversarial attacks. 

\subsection{Summary of Proposed Methodology}
In Algorithm \ref{alg:attack_method}, we summarize our framework of training ACGAN to generalize dynamic ransomware features and using a set of quality metrics to statistically evaluate the maliciousness of the generated samples.


\begin{algorithm}[h]
 \caption{Generate dynamic adversarial logs to bypass ransomware detector.}
\begin{algorithmic}
\State \textbf{Input}: Desired number of generated malicious samples $K$, quality threshold $\tau$, training data    
\State Step 1: Segmentation and dimension conversion. 
\State Step 2: Train AC-GAN. 
\State Step 3: Generate $G_m$ such that $|G_m| = K$. 
\State Step 4: Apply quality assessment procedure on $G_m$ as in Algorithm \ref{alg:quality_assessment}. 
\If{$K-K_0 = 0$} 
    \State Stop 
\Else 
    \State  Generate $G_m\textprime$ with $|G_m\textprime| = K-K_0$. Repeat until all generated segments pass quality assessment. 
\EndIf
\State Step 5: Concatenation. 
\State Step 6: Feed the logs into ransomware detectors.
\end{algorithmic}
\label{alg:attack_method}
\end{algorithm}
\section{Experiment Results} \label{sec:exp}

\subsection{Ransomware Classification on Real Data}
Machine learning can be efficient, scalable and accurate at recognizing malicious attacks. We first demonstrate its benefits for highly effective ransomware detection. The training and testing ratio is set at $80\%:20\%$, where the training set contains 1292 benign samples and 3736 malicious samples, and the test set contains 324 benign samples and 934 malicious samples. After feature mapping, each execution log is represented as a sequence of events, and the sequence length is set to be 3000, where shorter sequences are padded with zeros to ensure same length. 

We consider several popular classifiers including Text-CNN\cite{kim2014convolutional}, XGBoost \cite{chen2016xgboost}, linear discriminant analysis (LDA), Random Forest\cite{breiman2001random}, Naive Bayes\cite{mccallum1998comparison}, support vector machine with linear kernel (SVM-linear), and support vector machine with radial kernel (SVM-radial). For fair comparison, all classifiers are trained on the same sequences and no further feature extraction such as $n$-gram is performed prior to the classification algorithms. We note that the raw features are not $1$-gram modeling, which counts event occurrences. We report the classification accuracy, false positive rate (FPR), true positive rate (TPR), $F_1$-score 
and area under the ROC curve (AUC) 
for all selected classifiers. 



As seen in Table \ref{tab:classification_raw_features}, Text-CNN achieves the highest accuracy at 0.9890, low false positive rate at 0.03, highest true positive rate at 0.9989, highest F-score at 0.9796 and highest AUC at 0.9950 among all other selected classifiers. XGB performs second best with accuracy at 0.931 and lowest false positive rate at 0.023. All other classifiers either suffer from low accuracy or high false positive rate. However, we expect $n$-gram feature extraction will greatly improve the other classifiers' performance. 

Due to Text-CNN's superior performance, we naturally use it as a feature extractor via the last pooling layer and retrain all the other classifiers on the embedding extracted via Text-CNN. 
We observe significant improvement of other classifiers composed with Text-CNN, as seen in Table \ref{tab:ml_performance_textcnn_space}. 

\begin{table}[h]
\centering
\begin{tabular}{ | l | c | r |c|c|c|}
\hline
  Classifier & Accuracy & FPR & TPR & $F_1$-score  & AUC \\ \hline
  Text-CNN & 0.9890 & 0.030 &0.9989 & 0.9796& 0.9950 \\ 
  XGB & 0.9308 & 0.023 & 0.7963  &0.8557&0.8869\\
  LDA &  0.5048 & 0.574 & 0.7698   & 0.4077 & 0.6136 \\
   Random Forest   &0.9348 &0.213& 0.9861  &0.9497 &0.8866\\
Naive Bayes  & 0.8704 &0.250 & 0.9122  & 0.7488 &0.8457 \\

  SVM-linear & 0.4420 &0.074 & 0.3587  &0.4906 & 0.8130\\
  SVM-radial & 0.7417 &0.997& 0.9979 & 0.0061&0.9055\\ \hline

\end{tabular}
\caption[Performance table 1]{Classification performance on the test set. Text-CNN achieves the highest accuracy at 0.989 and low false positive rate at 0.03 among all selected classifiers. XGB performs second best with accuracy at 0.931 and lowest false positive rate at 0.023. All other classifiers either suffer from low accuracy or high false positive rate. }
\label{tab:classification_raw_features}
\end{table}

\begin{figure}
\centering
\includegraphics[width=0.5\textwidth]{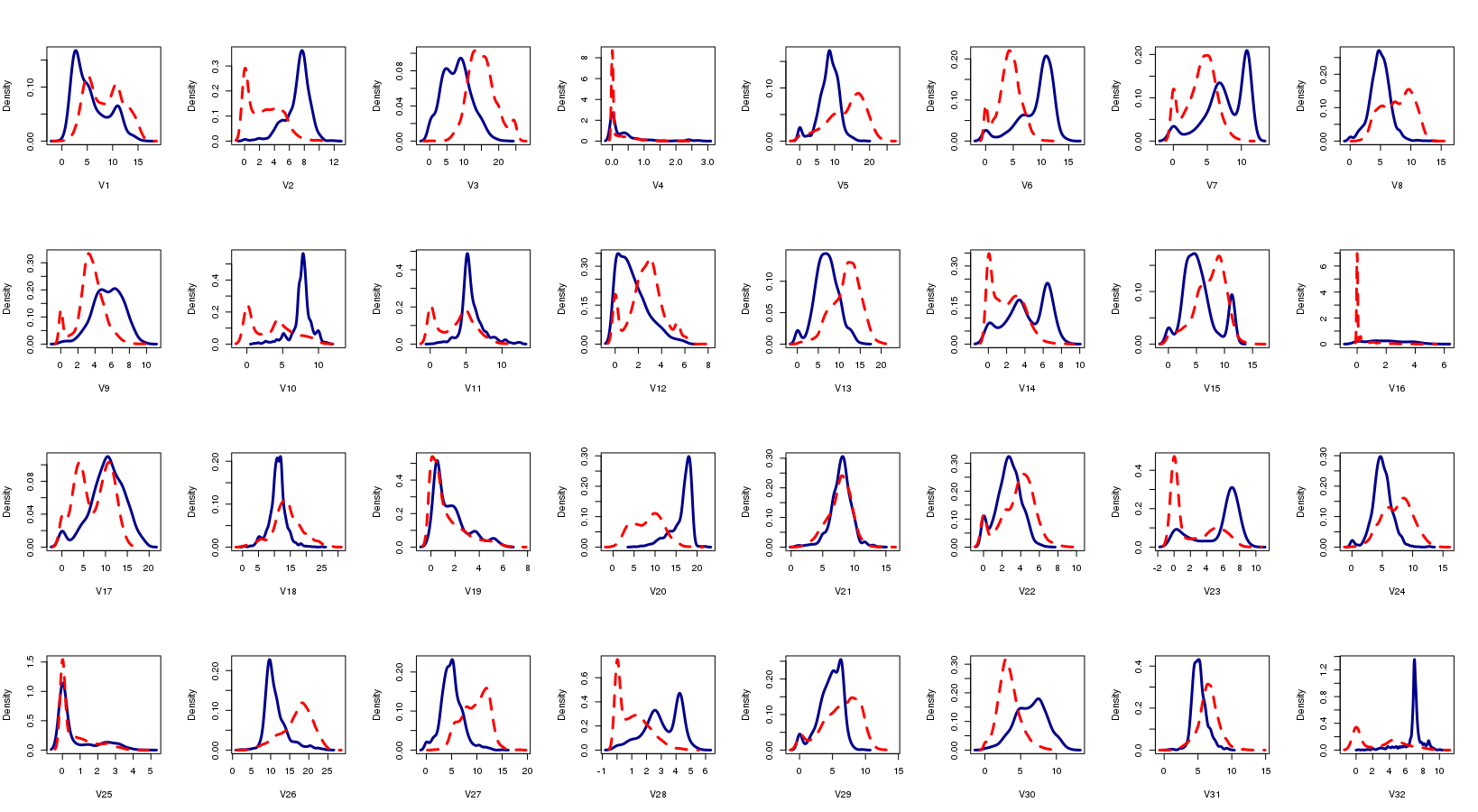}

\caption{\label{fig:density_plot} Class-conditional density plot for each dimension in Text-CNN feature space. Red denotes the malicious class and blue denotes the benign class. Text-CNN as a feature extractor helps separate the samples from two classes, as indicated by the density plots. The features extracted from Text-CNN are in $\mathbb{R}^{32}$.}
\end{figure}

\begin{table}[h]
\centering
\resizebox{\columnwidth}{!}{%

\begin{tabular}{ | l | c | r |c|c|c|}
\hline
  Classifier & Accuracy & FPR & TPR  & F-score  & AUC \\ \hline

 XGB $\circ$  Text-CNN & 0.9841 & 0.0032& 0.9475 &0.9685 &0.9722\\
 LDA  $\circ$ Text-CNN   &  0.9865 &0.0494& 0.9989  & 0.9731 &0.9977\\
 Random Forest  $\circ$ Text-CNN   &0.9833  & 0.0556&  0.9968  &0.9497 &0.9706\\
Naive Bayes  $\circ$ Text-CNN & 0.9666 &0.1111 & 0.9936 & 0.9320 &0.9906 \\
 SVM-linear  $\circ$ Text-CNN   &0.9881  & 0.0432&  0.9989 &0.9764 &0.9974\\
  SVM-radial $\circ$  Text-CNN   & 0.9897  & 0.0228& 0.9957 &0.9797 &0.9993\\ \hline

\end{tabular}}
\caption[Performance table 1]{Classification results on the test set. All the classical classifiers performance improve significantly using Text CNN as a feature extractor.}
\label{tab:ml_performance_textcnn_space}
\end{table}

\begin{figure}
\centering
\includegraphics[width=0.15\textwidth]{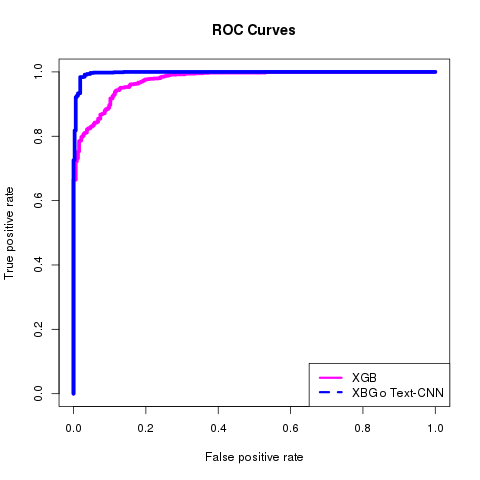}
\includegraphics[width=0.15\textwidth]{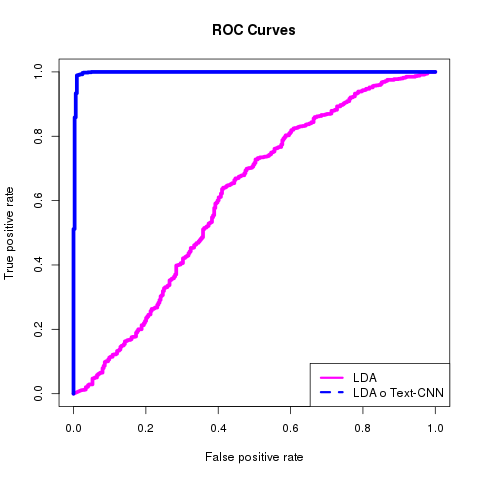}
\includegraphics[width=0.15\textwidth]{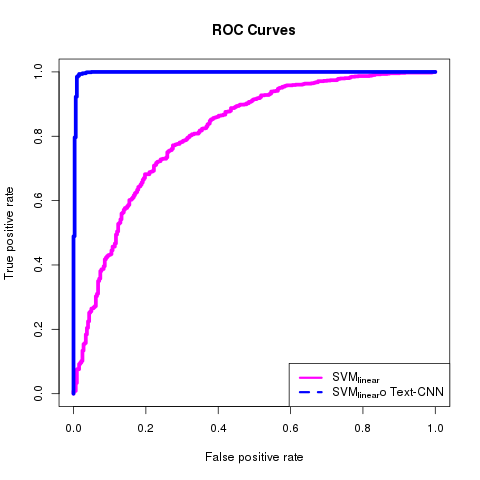}
\caption{\label{fig:roc} ROC curves of XGB, LDA, SVM compared with XGB$\circ$  Text-CNN, LDA$\circ$  Text-CNN and SVM$\circ$  Text-CNN. When using Text-CNN as a feature extractor and retraining XGB, LDA, SVM in the Text-CNN embedding subspace, all the composed classifiers possess significantly higher classification efficacy measured by AUC, $F_1$-score, accuracy FPR and TPR. }
\end{figure}

It is only worthwhile to evaluate the resiliency of a highly effective ransomware classifier. Based on Table \ref{tab:ml_performance_textcnn_space}, the high effective ransomware classifiers are Text-CNN, XGB$\circ$  Text-CNN, LDA$\circ$  Text-CNN,  Random Forest  $\circ$ Text-CNN, Naive Bayes  $\circ$ Text-CNN and SVM$\circ$  Text-CNN. In our experiment results, Text-CNN, whether as a classifier on its own or as a feature extractor, is most likely to be selected by a security defender. Although knowledge of the defender's ransomware classifier is not needed by our analysis methodology, we evaluate the adversarial detection rate against Text-CNN based classifiers. 


\subsection{Generate Adversarial Segments}
We follow the steps in Section \ref{subsec:acgan} to train an AC-GAN \cite{odena2016conditional}, where we set the batch size to be 100, the latent dimension to be 100, and the training is stopped at the 80-th epoch. 
After training, we obtain 5029 segments from the malicious class $Y=1$. We round the segments to the nearest integer and denote this set as $G_m$. 



\subsection{Adversarial Quality Assessment}
\begin{figure}
    \centering
    \includegraphics[width=0.15\textwidth]{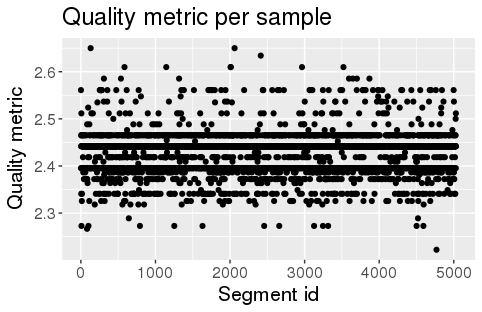}
    \includegraphics[width=0.15\textwidth]{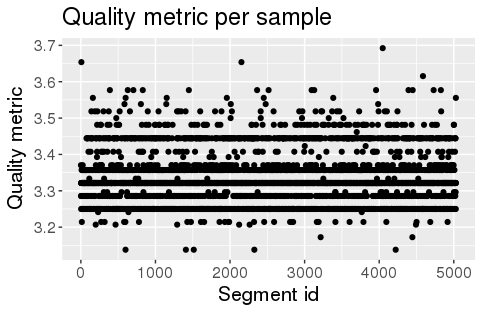}
    \includegraphics[width=0.15\textwidth]{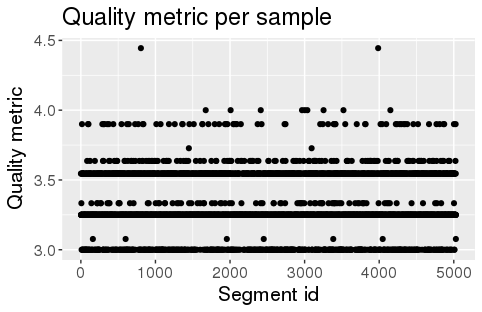}
    \caption{Adversarial quality metric $q_{te}$ for 4-,5-,6-grams. All the generated segments have $q_i \geq \tau$, where $\tau = 1.5$ and $\min\{q_i\} = 1.9$. Hence the generated segments have minimum of almost twice the unique malicious signatures than the unique benign signatures for 4-,5-,6-grams.  }\label{fig:sample_quality_metric}
\end{figure}

A successful evasion means the generated malicious samples not only fool ransomware classifier, but also persists maliciousness based on certain metrics. Following Section \ref{subsec:quality_assess}, we compute the adversarial quality metric $q$ of each GAN-generated sample for $n$-grams with $n\in \{3,4,\dots, 7\}$. 
Figure \ref{fig:sample_quality_metric} shows the quality metric in $y$-axis against each generated segment in $x$-axis for 4-, 5-, 6-grams. We set the quality threshold to be $\tau = 1.5$, which means a qualified generated segment with statistically measured maliciousness would need to match over 50\% of the unique malicious $n$-grams than the unique benign $n$-grams.

\begin{figure}
\centering
\includegraphics[width=2.2in, height=1.4in]{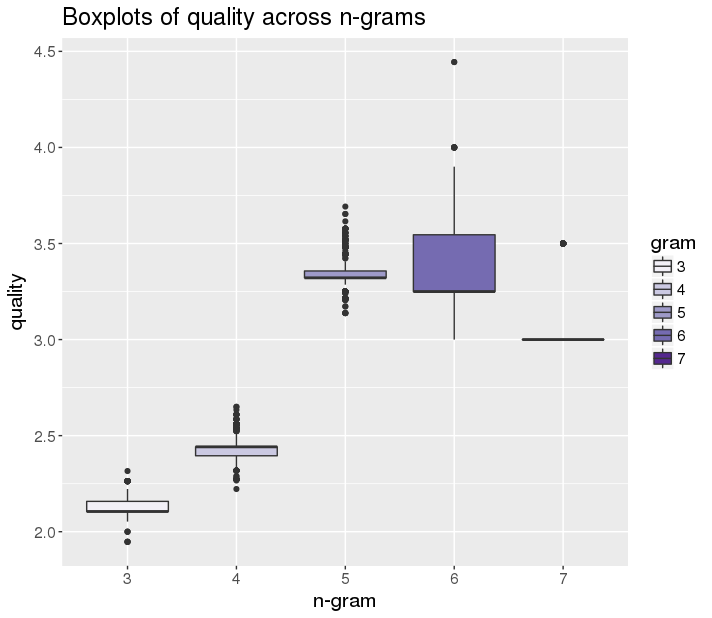}
\caption{\label{fig:boxplot} Boxplots of $Q$ to evaluate the generated batch quality. All the generated segments have $q_i \geq \tau$, with $\min\{q_i\} = 1.9$ for all $n\in \{3, 4, \dots, 7\}$-grams. }
\end{figure}

We also plot the batch-based quality metric $Q$ for $n = \{3, 4, \dots, 7\}$-grams, as represented in boxplots in Figure \ref{fig:boxplot}. As shown in the boxplots, all the generated segments are statistically much closer to the real malicious class with $q_i \geq \tau$ and $\min\{q_i\} = 1.9$.

All the generated and qualified segments are concatenated randomly to produce 1257 
execution logs. 

\subsection{Evasion}
The highly performing ransomware detectors Text-CNN, XGB $\circ$ Text-CNN, LDA $\circ$ Text-CNN, Random forest  $\circ$ Text-CNN, Naive Bayes $\circ$ Text-CNN, SVM-linear $\circ$ Text-CNN, SVM-radial $\circ$ Text-CNN are applied on the adversary-generated logs. We report the number of detected samples and the detection rate in Table \ref{tab:attack_eval}.

Most of the classifiers significantly degrade in detection performance, where Text-CNN, LDA $\circ$ Text-CNN,  Naive Bayes $\circ$ Text-CNN, SVM-linear $\circ$ Text-CNN fail to detect any generated malicious samples, while XGB $\circ$ Text-CNN detects 12.73\% correctly and  Random forest  $\circ$ Text-CNN detects 36.35\% correctly. The most robust classifier turns out to be SVM-radial $\circ$ Text-CNN in this experiment with 100\% detection rate. This can be due to its nonlinear boundary in the Text-CNN latent feature space. However only one classifier out of all seven highly effective classifiers is resilient to our bypass scheme. Our adversarial detection result clearly indicates that this is a potential vulnerability for ML-based ransomware detection systems.



\begin{table}[h]
\center
\begin{tabular}{| l | c | c|}
\hline
  Classifier &  No. detected  & Detection rate (\%) \\\hline
   Text-CNN & 0 & 0 \\
  XGB $\circ$  Text-CNN & 16 & 12.73\\
  LDA $\circ$  Text-CNN & 0 & 0  \\
    Random forest $\circ$  Text-CNN & 457 & 36.35  \\
 Naive Bayes $\circ$  Text-CNN & 0 & 0  \\
  SVM-linear$\circ$  Text-CNN  & 0 & 0 \\

  SVM-radial$\circ$  Text-CNN  & 1257&  100\%\\  \hline
\end{tabular}
\caption[Performance table 1]{Adversarial detection rate on the generated malicious samples. Six of the seven highly effective classifiers degrade severely in performance and only one classifier persists resiliency against attacks. This quantifies the attack surface for these ML-based ransomware detection algorithms. The non-linear boundary of SVM-radial$\circ$  Text-CNN effectively detects the adversarial samples.}
\label{tab:attack_eval}
\end{table}

\subsection{Latent Feature Space Investigation}

We investigate why most of the highly effective classifiers fail to predict the adversarially generated samples correctly. We use the last pooling layer from Text-CNN as a feature extractor and refer to the space of features extracted by Text-CNN as the latent feature subspace.
The classifiers that achieve effective and competitive classification performance are XGB, LDA, Random Forest, Naive Bayes and SVM trained in the latent feature subspace. Text-CNN the classifier itself has linear boundaries via the fully connected layer in the latent feature subspace. 
Hence one natural investigation starts at how the generated samples and the real samples relate in the latent feature subspace induced by Text-CNN, in comparison with their relationship in the original feature space, consisting of the raw execution logs.  

\begin{figure}
\centering
\includegraphics[width=0.15\textwidth]{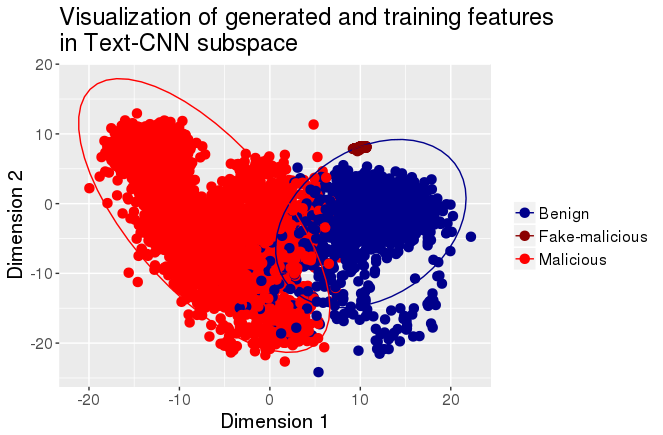}
\includegraphics[width=0.15\textwidth]{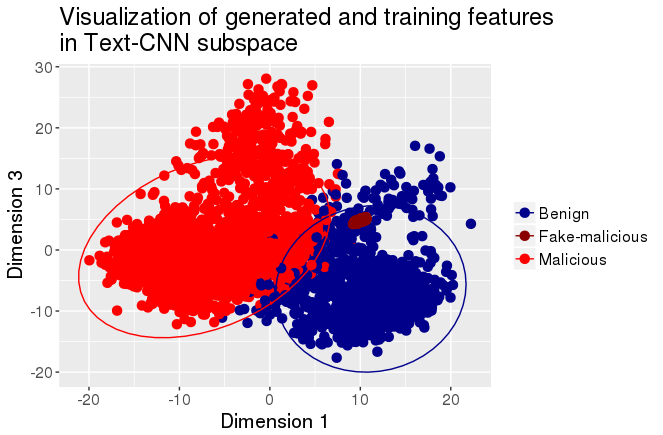}
\includegraphics[width=0.15\textwidth]{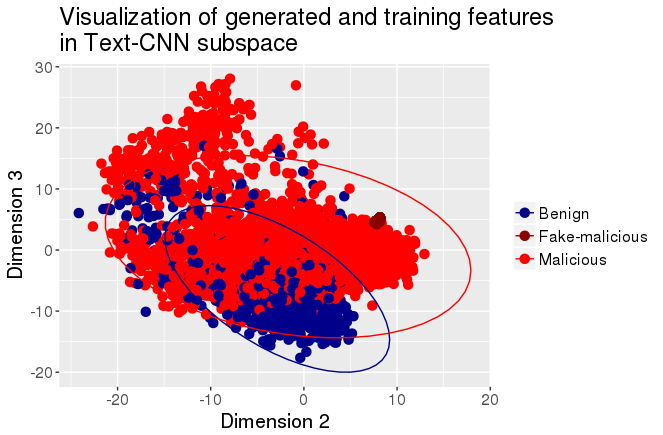}

\caption{\label{fig:subspace_tsne} Visualization of the Text-CNN extracted features for (left) PC-dimension 1 vs PC-dimension; (middle) PC-dimension 1 vs PC-dimension 3; (right) PC-dimension 2 vs PC-dimension 3. The generated malicious samples are colored in dark red, and lie closer to the benign set in Text-CNN subspace. We draw the 95\% data eclipse around the scattered points.}

\end{figure}

Represented in 2-D visualization, Figure \ref{fig:subspace_tsne} shows that the generated samples, in dark red, lie close to a linear boundary but much closer to the real benign samples in the Text-CNN latent feature subspace. However as shown in Section \ref{sec:exp}, most of the generated samples match more than twice of the unique ransomware signatures than the unique benign signatures. This motivates us to explore the $L_2$ distance between the real malicious samples and real benign samples, as well as between the generated samples and the real samples in both the latent feature subspace and the original feature space. 

Denote the latent features of the generated malicious logs as $F_g$, the latent features of the training malicious logs as $F_{tr,m}$ and the latent features of the training benign logs as $F_{tr, b}$. Similarly, for the test data, the latent malicious and benign features are denoted as $F_{te, m}$ and $F_{te,b}$ respectively.

\begin{figure}
\centering
\includegraphics[width=0.23\textwidth]{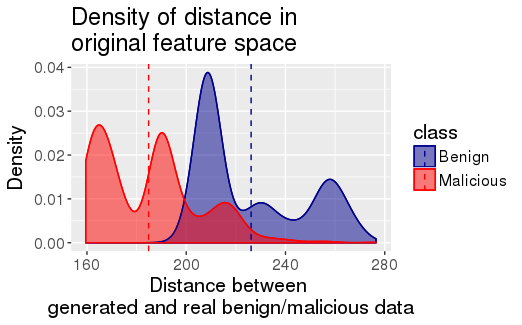}
\includegraphics[width=0.23\textwidth]{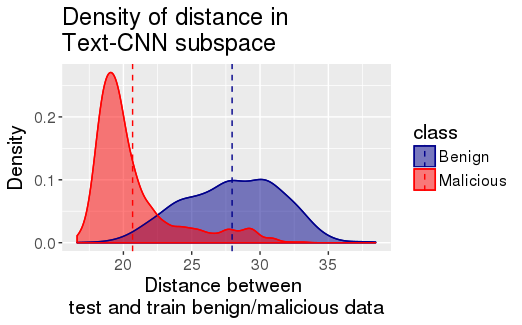}
\caption{\label{fig:density_test_train} Density plot of the distances between real benign and real malicious logs in both original feature space and Text-CNN latent feature space. }
\end{figure}

We plot the density of the $L_2$-distances between test malicious data and training data, both of which are real samples. The left figure in Figure \ref{fig:density_test_train} shows, in the original feature space, the density of the $L_2$ distance $D_{tr, te, m}$ between the malicious test logs and the training malicious logs in red and the density of the $L_2$ distance $D_{tr, te, b}$ between the malicious test logs and the training benign logs in blue. The dashed red and blue vertical lines represent the means of $D_{tr, te, m}$ and $D_{tr, te, b}$ respectively. On average, the malicious test logs are closer to the training malicious logs than to the training benign logs. However in the original data space, the distributions of distances are not very well-separated and this is also reflected in the algorithm performance on the original data space as shown in Table \ref{tab:classification_raw_features}.

The right figure in Figure \ref{fig:density_test_train} plots the density of the $L_2$ distance $d_{tr, te, m}$ between $F_{te, m}$ and $F_{tr, m}$ in red and the density of the $L_2$ distance $d_{tr, te, b}$ between $F_{te, m}$ and $F_{tr, b}$ in blue. The dashed red and blue vertical lines represent the means of $d_{tr, te, m}$ and $d_{tr, te, b}$ respectively. $F_{te,m}$ is much closer to $F_{tr, m}$ than to $F_{tr, b}$. The distances are consistent across original feature space and the latent feature subspace. This observation is expected since the malicious samples should be close together in either feature space. 



\begin{figure}
\centering
\includegraphics[width=0.23\textwidth]{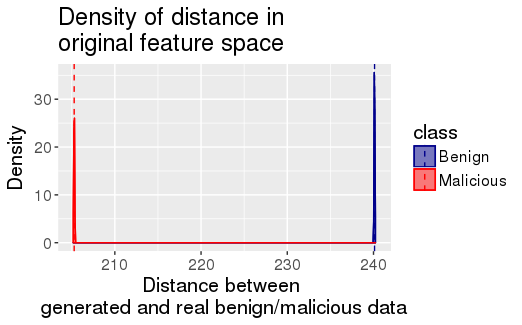}
\includegraphics[width=0.23\textwidth]{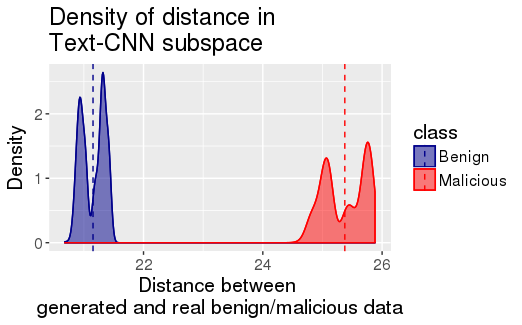}
\caption{Density plot of the distances between generated logs and real logs in both original feature space and Text-CNN latent feature space.  }
\label{fig:density_plot_gan_train}
\end{figure}

Next we investigate whether the observed phenomenon extends to the generated samples and real samples. The left figure in Figure \ref{fig:density_plot_gan_train} plots, in the original feature space, the density of the $L_2$-distance $D_{tr, g,m}$ between the generated logs and the training malicious logs in red and the density of the $L_2$ distance $D_{tr, g,b}$ between the generated logs and the training benign logs in blue. The dashed red and blue vertical lines represent the means of $D_{tr, g,m}$ and $D_{tr, g,b}$ respectively. The generated malicious logs are much closer to the real malicious logs than to the real benign logs in the original feature space.


The right figure in Figure \ref{fig:density_plot_gan_train} plots, in the latent feature space, the density of the $L_2$-distance $D_{tr, g,m}$ between $F_g$ and $F_{tr, m}$ in red and the density of the $L_2$ distance $D_{tr, g,b}$ between $F_g$ and $F_{tr, b}$ in blue. The dashed red and blue vertical lines represent the means of $D_{tr, g,m}$ and $D_{tr, g,b}$ respectively. $F_g$ is much closer to $F_{tr, b}$ than to $F_{tr, m}$. Figure \ref{fig:density_plot_gan_train} shows that in the Text-CNN feature subspace, the generated logs are closer to the benign logs, while in the original feature space, the generated logs are closer to the malicious logs. This phenomenon indicates that the generated adversarial samples lie in the blind spot of the Text-CNN algorithm.






\section{Deployment Opportunity} \label{sec:deployment}

\begin{figure}[t!]
\centering
\includegraphics[width=0.4\textwidth]{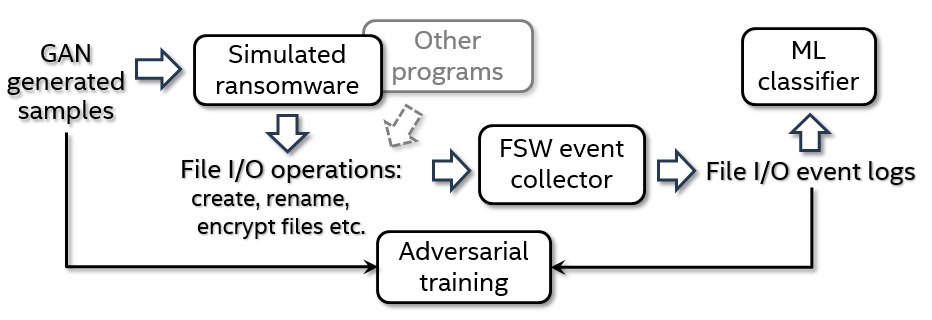}
\caption{\label{fig:simulation} The flow to synthesize actual samples based on GAN-generated I/O sequences. The goal to deploy our proposed system is to use the generated samples for adversarial training and improve model resiliency.}
\end{figure}

We develop a ransomware simulation program to demonstrate how to generate actual malicious file I/O operations based on the GAN generated feature sequences, as seen in Figure ~\ref{fig:simulation}. When the program sees feature sequences such as “File rename”, “File created” and “File change with high entropy”, it will rename an arbitrary file, create the other arbitrary file and then encrypt a victim file. The I/O events collected when executing the simulated ransomware is very close to the GAN generated feature sequence, such that the simulated ransomware can bypass ML detection while maintaining malicious encryption behaviors. We make sure the I/O operations are completed in a short time window to minimize noisy I/O events added from other benign processes. 
\section{Discussion} \label{sec:discuss}





 
We describe a framework via generative adversarial network to synthesize dynamic ransomware samples and propose a set of adversarial quality metrics via statistical similarity to quantify the maliciousness of the GAN-generated samples. We demonstrate that six of the seven highly effective ransomware classifiers fail to detect most of the GAN-generated samples.  

In our next steps, we will continue developing an automatic tool for monitoring applications and harvesting more benign logs. The addition of benign samples can augment our training set to better reflect the practical scenarios of more benign-ware than malware.  
We also plan to test the real-world efficacy of our proposed adversarial log generation system against the machine learning based anti-malware solutions in the market. Additionally, while we hope to extend our adversarial generation framework to semi-supervised or unsupervised malware classification tasks. 

the unsupervised ransomware detection algorithms were not considered in this paper. As many unsupervised machine learning algorithms have been developed, how to properly access their resiliency and trustworthiness is also important. In 


Our proposed framework should be utilized as a defense capability for developing a resilient model for detecting ransomware in the field. As described in Section \ref{subsec:eval}, a defender can use the adversarial detection rate as a metric to quantify the resilience of the ransomware detector against adversarial attacks. The defender can use the GAN-generated samples as part of the training procedure to update the defender's classifier.


Our proposed quality assessment approach can be leveraged even when the model is deployed and is in use in the field to track the changes in distance between generated and real samples. These robustness mechanisms must be considered as an integral part of an adversary-resilient malware classifier. 

Our case study for evaluating a broad range of ransomware classifiers also demonstrates the pitfalls in selecting classifiers based on high accuracy and low false-positives which is typical today in malware detection. After a deeper analysis of generating quality adversarial samples, the most robust classifier is verified to be SVM-radial$\circ$Text-CNN in our experiment. This analysis may form the basis of selecting multi-classifier ensemble-based approaches to act as a defense-in-depth against adversarial probing attacks once the ransomware classifiers are deployed in the field. In our specific case study, a weighted score between the XGB$\circ$Text-CNN classifier and the SVM-radial$\circ$Text-CNN classifier gives the defender much more coverage in the space of execution logs for ransomware.

It is important to note that our framework is still useful to enforce the resiliency of the ransomware detection model even when the model is deployed on a platform using software and hardware-based Trusted Execution Environments (TEEs) that protect the run-time confidentiality and integrity of the classifier(s) while in-use - providing the defender with an additional tool to continue to enforce the security objectives consistently even post the training stages.



\bibliographystyle{plain}
\bibliography{sample}

\end{document}